\pdfoutput=1

\documentclass[11pt]{article}

\usepackage{acl}

\usepackage{times}
\usepackage{latexsym}

\usepackage[T1]{fontenc}

\usepackage[utf8]{inputenc}

\usepackage{microtype}

\usepackage{inconsolata}

\usepackage{graphicx}

\usepackage{multirow}

\usepackage{lipsum}
\usepackage{xurl}
\usepackage{hyperref}
\usepackage{todonotes}
\usepackage{duckuments}
\usepackage{amsmath}

\renewcommand\paragraph[1]{
\vspace{0.15cm}
\noindent 
\textbf{#1}
}

\title{\textit{Is It Really Long Context if All You Need Is Retrieval?} \\ Towards Genuinely Difficult Long Context NLP}

\author{Omer Goldman\thanks{Equal contribution}, \ Alon Jacovi\footnotemark[1], \ Aviv Slobodkin\footnotemark[1], \\
\bf Aviya Maimon\footnotemark[1], \ Ido Dagan, \  Reut Tsarfaty \\
Bar-Ilan University \\
\texttt{omer.goldman@gmail.com}
}

\begin{document}
\maketitle

\begin{abstract}

Improvements in language models' capabilities have pushed their applications towards longer contexts, making long-context evaluation and development an active research area.
However, many disparate use cases are grouped together under the umbrella term of ``long-context'', defined simply by the total length of the model's input, including -- for example -- Needle-in-a-Haystack tasks, book summarization, and information aggregation. 
Given their varied difficulty, 
in this position paper we argue that conflating different tasks by their context length is unproductive. As a community, we require a more precise vocabulary to understand what makes long-context tasks similar or different.
We propose to unpack the taxonomy of long-context based on the \textit{properties that make them more difficult} with longer contexts. 
We propose two orthogonal axes of difficulty: (I) \textit{Dispersion:} How hard is it to find the necessary information in the context? (II) \textit{Scope:} How much necessary information is there to find?
We survey the literature on long context, provide justification for this taxonomy as an informative descriptor, and situate the literature with respect to it.
We conclude that the most difficult and interesting settings, whose necessary information is very long and highly dispersed within the input, is severely under-explored.
By using a descriptive vocabulary and discussing the relevant properties of difficulty in long context, we can implement more informed research in this area.
We call for a careful design of tasks and benchmarks with \textit{distinctly} long context, taking into account the characteristics that make it qualitatively different from shorter context.

\end{abstract}

\section{Introduction}

\begin{figure}[t]
    \centering
    \includegraphics[width=0.99\columnwidth]{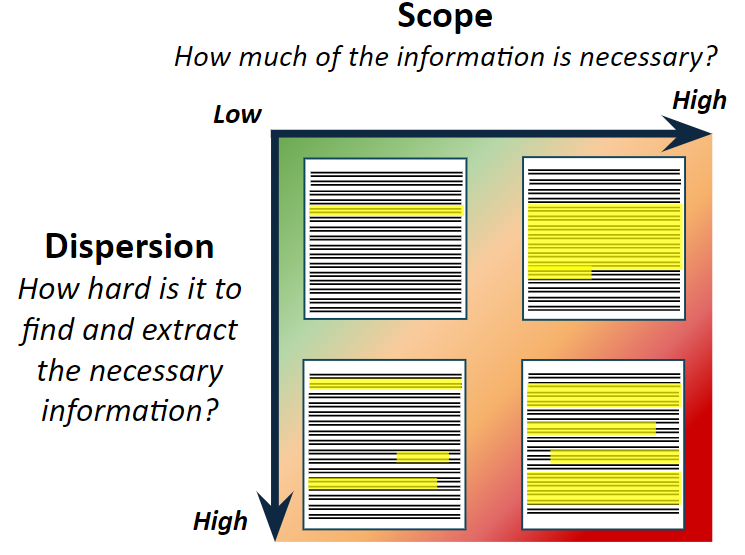}
    \caption{A taxonomy of long context tasks based on the distribution of the needed information in the text. Tasks with larger scope and higher dispersion are more difficult (indicated by shade) and more indicative of the long context capabilities of large language models.}
    \label{fig:1st_page}
\end{figure}

The ability to deal with ever-longer contexts has been one of the most notable trends among the emerging capabilities of large language models (LLMs). Starting with a few hundred tokens as the maximal input length of the first attention-based LLMs \cite{devlin-etal-2019-bert, raffel2020exploring}, contemporary models are -- \textit{technically} -- able to process up to 128k and even 1M tokens \citep{geminiteam2024gemini, openai2024gpt4}. %

The demand to evaluate LLMs in this setting has led to a line of research on designing long-context tasks and benchmarks, in order to systematically understand models' capabilities and drive their development.
However, the field has generally a sole recurring descriptor to define such measurements by -- simply, the length of the context.
For example, long-context benchmarks group tasks mostly by length in words \citep[e.g.,][]{shaham-etal-2022-scrolls, bai2023longbench, zhang2024inftybench}. %
This leads to qualitatively different measurements being conflated together, with conclusions about long-context capabilities being extended from one class of tasks to others. 
The community is, of course, aware that, for example, tasks which require a small part of the input are different from tasks that require a large part of it. But we ask the more general question: What are the properties that differentiate tasks when conditioned on their context length? What can we accomplish with such a distinction?

In this position paper, we claim that the current landscape of works on long-context evaluation will greatly benefit from a more fine-grained characterization of long-context task design. We argue that judging LLMs by their ability to process long sequences, while disregarding the task they process them for, overlooks the characteristics that make longer inputs more difficult, and interesting to research, to begin with~(\S\ref{sec:survey}).

For example, Needle in a Haystack tasks~(NIAH; \citealp{ivgi-etal-2023-efficient,mohtashami2023landmark}) involve queries whose main challenge is finding the relevant information in a long context, without requiring much further processing. Synthetic NIAH datasets are, of course, easier than their natural equivalents \cite{ivgi-etal-2023-efficient}, but the ``natural vs. artificial'' classification is not informative in our setting, since it applies equally for tasks regardless of context length. What, then, is an informative property? What makes long-context tasks different from each other? For example, multiple-needle variants of NIAH \cite{hsieh2024ruler}, or those that position the ``needles'' closer or farther apart \cite{levy2024task}. Evidently, ``the number of tokens in the input'' is not a sufficient descriptor. 

To resolve this roadblock, we present a taxonomy of long-context tasks for the different factors that make them harder \textit{when controlling for context length}~(\S\ref{sec:taxonomy}). This taxonomy is derived by surveying the long-context literature and surfacing the most salient points of distinction between various tasks. 
We focus on (I) how difficult it is to find and extract the required information from the input (its \textit{dispersion} in the input), and (II) the absolute quantity of required information to solve the task (its \textit{scope}). See  \autoref{fig:1st_page} for a summary.

To understand this categorization and its utility, we review the literature on long-context evaluation and position the works with respect to those factors. We find that the most challenging setting, in which a large quantity of required information is present in a dispersed manner that is difficult to extract, is significantly under-explored~(\S\ref{sec:classification}).

Finally, acknowledging the inherent and legitimate reasons behind the focus on context length as the sole descriptor of difficulty, we discuss possible paths forward for designing more reliable measurements of long-context capabilities when utilizing a more nuanced vocabulary~(\S\ref{sec:next}).

\section{Task Design in Long Context}
\label{sec:survey}

Evaluating the performance of NLP models over very long contexts is a fast-changing area of research~\cite{bishop2024longdocfactscore, wu2024long}. Measurements are regularly updated to account for new capabilities which improve with extrapolation architectures~\cite{Vaswani2017AttentionIA,su2024roformer} and training data~\cite{he2023lost,Chen2023LongLoRAEF}. Evaluators were tasked with designing measurements of long-context capabilities cheaply, efficiently, and quickly, while matching realistic use cases as much as possible. The most common way of differentiating long-context tasks, besides the context's length, is whether they are naturally-constructed or synthetically-constructed~\cite{tay2020long, bai2023longbench, hsieh2024ruler}.

\paragraph{Natural construction.} 
A simple yet effective way of ``moving the goalpost'' for context length is by modeling long-context tasks based on short-context tasks. 
This was done, for example, with 
QA (\citealp{narrativeQA}, cf. \citealp{dunn2017searchqa}), summarization (\citealp{huang-etal-2021-efficient}, cf. \citealp{narayan-etal-2018-dont}), and NLI (\citealp{koreeda-manning-2021-contractnli-dataset}, cf. \citealp{williams-etal-2018-broad}). Specialized domains like legal \cite{bruno2022lawngnli, nguyen2024captain} and literature \citep{wang-etal-2022-squality, kryscinski-etal-2022-booksum} often involve longer texts, turning typically short-context tasks such as QA and NLI into long-context scenarios.
Another more native methodology is to create new tasks which inherently require a long context,
such as multi-document summarization \cite{fabbri2019multinews, angelidis-etal-2021-extractive}, survey generation \cite{gao2024large}, and structured data aggregation \cite{caciularu2024tact}.
Both methodologies share the constraint that, due to their natural construction (i.e., using natural text), once created, they are difficult to modify for longer contexts as models' long-context capabilities improve.

\paragraph{Synthetic construction.} 
A more flexible approach, sacrificing natural construction for length control, is to use distractors to synthetically increase the context length. 
This method allows for cheap and efficient (in terms of task construction cost) evaluation of models' full context length capabilities, with difficulty adjusted by controlling the distractors.
Tasks like Needle-in-a-Haystack \citep[NIAH;][]{ivgi-etal-2023-efficient,kamradt2023needle} and PassKey retrieval \citep{mohtashami2023landmark} were created to evaluate a model’s ability to pinpoint specific information amid lengthy distractors. 
Flexible and effective against existing models, they became standard benchmarks for evaluating new long-context models \citep{glm4, jiang2024mixtral}.
Followup studies have complicated these tasks by increasing the number of critical details to locate \citep{arora2023zoology, liu2024world} and changing their position within the input \cite{liu2024lost, levy2024task}.

\paragraph{Limitations of the status quo.}
NIAH-like tasks aim to assess information retrieval capabilities, yet 
many ``naturally constructed'' QA and reading-comprehension tasks 
with trivial questions about a long context accomplish the same goal. At the same time, ``multiple needles'' NIAH can increase difficulty not by increasing the quantity of needles or length of input, but by \textit{adding distractors} between needles~\cite{levy2024task}. What can systematically explain the different variables at play, in order to inform better task design in the future?

Clearly, there are \textit{underlying qualitative differences} that discriminate between these various tasks besides their natural and synthetic constructions, and besides their actual context length. 
Therefore, we require a more informative vocabulary to discuss the goals of each task design, what it accomplishes, and what it does not, in terms of measuring long-context capabilities.

\section{What Makes Long Context More than Retrieval?}
\label{sec:taxonomy}

We require a taxonomy to capture task difficulty variations beyond mere ``number of tokens''.
We focus on the information that is canonically \textit{required} to solve the task as the conditioning variable. %
Our classification can be summarized via the following two questions, when asked about a given task:

\vspace{0.05cm}
\noindent (I) \textit{How difficult is it to find and extract the required information?}

\vspace{0.05cm}
\noindent (II) \textit{How much information is needed to be found?}
\vspace{0.05cm}

\noindent Assuming that some highlighting of the relevant information is needed to solve the task (see \autoref{fig:1st_page}), the latter question asks how much text is highlighted, while the former addresses the challenge of locating the relevant text for highlighting.

For instance, consider the task of summarizing a book, in comparison to a NIAH task of identifying a numerical detail in a long financial report (e.g., ``how much did the company earn in 2015?''). 
Although both tasks involve long texts, the \textit{information required} and its \textit{accessibility} vary significantly. 
The NIAH task focuses on localized, identifiable information, while summarization requires extracting key details dispersed throughout the text, tangled together with irrelevant content. Therefore, we can say that the book summarization task is more difficult on both axes (I) and (II). 

Below we give more formal descriptions of the two axes characterized by the questions above.

\paragraph{(I) Dispersion.} 
Although the question above intuitively defines ``difficulty of information finding'',
 we offer a more concrete description. Between two similar tasks, we consider the information harder to find in one task compared to another if:
(1) it is more obscured (e.g., linguistically, semantically,  contextually, etc); (2) it is more sparse, such that it is interspersed with non-required information; (3) its indicators are less redundant, such that there are fewer places in the document where the same information is available.

\paragraph{(II) Scope.}
The property of scope is simpler, and refers to the minimal quantity of information needed to solve the task. Importantly, we are not concerned with precise metric for ``quantity of information'' at this stage -- it can refer to quantity of tokens, sentences, relations, cells in a table, etc. Any metric that reliably captures difficulty for an established solver is sufficient for our purposes.

\paragraph{Illustrative example.} 
To illustrate, consider the Wikipedia entry for \textit{New York City} and a simple question: ``What is the estimated population of the city?'' Since the answer needs a small snippet of information, we say that the task has \textit{small scope}. And since it is easily accessible, we say that it has \textit{low dispersion}. Consider, instead, the question ``how many syllables are in this document?'' -- since this question requires the entire document to answer, we say that it has \textit{large scope}, but if we consider counting syllables as straightforward, then we say its \textit{dispersion} is still \textit{low}. Finally, with the question ``Was the city's mayor elected before or after the city was affected by Hurricane Sandy?'' -- since it requires snippets from at least two different areas of the text, we can say that when compared to the question about the city's population, the \textit{dispersion} is \textit{higher}, but not as high as for the question ``What makes the city a prominent place on the world stage?'' which poses a challenge on both axes.

\begin{figure}[t]
    \centering
    \includegraphics[width=0.99\columnwidth]{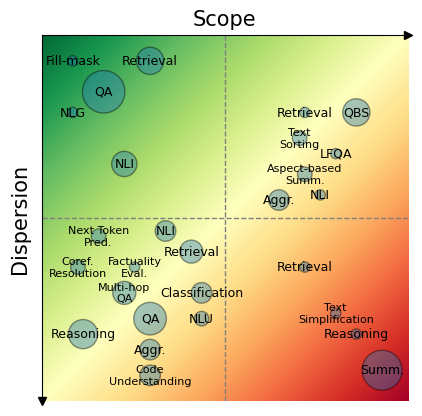}
    \caption{This figure illustrates our subjective judgment on the distribution of long-context benchmarks for each task, categorized by their scope and dispersion characteristics, with the four quadrants being marked by the dashed lines. Difficulty is expressed by shade, where \textcolor[rgb]{0.8,0.0,0.0}{red} is more difficult and \textcolor[rgb]{0.0,0.5,0.0}{green} in easier. Notably, some tasks, like Question-answering (QA), appear in multiple quadrants, as different benchmarks demand varying levels of scope and dispersion (e.g., a single fact versus multiple facts spread across a document). For a detailed breakdown of benchmarks and their task associations, refer to \autoref{sec:benchmark_scope_dispersion_classification}.}
    \label{fig:scatter_plot}
\end{figure}

\section{Challenging Long Context Is Under-Explored}
\label{sec:classification}

Revisiting the works surveyed in \S\ref{sec:survey}, they clearly differ with respect to both scope and dispersion.

\paragraph{With respect to \textit{dispersion}.} 
The information needed for tasks ranges from easily accessible to highly dispersed and difficult to detect.
On low dispersion we have NIAH \cite{kamradt2023needle, mohtashami2023landmark} and a myriad of factual single-hop QA datasets \cite[inter alia]{tseng2016towards, kočiský2017narrativeqa, kwiatkowski-etal-2019-natural, dasigi-etal-2021-dataset} in which the answer is relatively accessible.
Adding more snippets of information separated by distractors, either in the form of several needles \cite{arora2023zoology, hsieh2024ruler} or of hops in a multi-hop question \cite{trivedi2022musique, zhao-etal-2022-multihiertt}, complicates the information detection due to the need to find at least two snippets \cite{levy2024task}, thereby increasing dispersion. 
Dispersion can also be increased by making the detection of the information less straightforward
\cite[e.g.,][]{pang-etal-2022-quality} or requiring aggregation \cite{shaham-etal-2023-zeroscrolls}.
Lastly, summarization tasks are of a very high dispersion \cite{huang-etal-2021-efficient, wang-etal-2022-squality}, as they require the non-trivial detection of salient facts that are interwoven with the irrelevant text. %

\paragraph{With respect to \textit{scope}.} 
Tasks overwhelmingly target relatively small scope. In addition to the aforementioned NIAH tasks and their variants, many
QA datasets apply as well~\cite[inter alia]{li2023loogle, Zhao2023DocMathEvalEN, reddy2024docfinqa}.
A somewhat higher scope is achieved by datasets for query-based summarization \cite{zhong-etal-2021-qmsum, wang-etal-2022-squality}, %
and QA datasets with more obfuscated answers that require reading the text surrounding the answer for its verification~\cite{an2023leval, he2023lost}.
Although much higher on the scope ladder, book summarization is still limited in its scope as datasets include texts that are only of up to 20$k$ tokens \cite{huang-etal-2021-efficient, chen-etal-2022-summscreen, shaham-etal-2023-zeroscrolls}. 
Currently, tasks with the highest scope, requiring information across the entire input length, are artificial and of low dispersion, like common words extraction \cite{hsieh2024ruler}.

\paragraph{Conclusion.} 
\autoref{fig:scatter_plot} summarizes the above classification of tasks and datasets. Note that without a concrete definition of dispersion and scope, the plot is only an illustration that involves a good deal of subjective judgements. 
However, we conclude that (1) the majority of tasks designed to challenge LLMs in a long-context setting target either scope \textit{or} dispersion, 
such that (2) tasks that push current models' capabilities on \textit{both} axes are under-represented in the current landscape.

\section{Discussion: Towards Genuinely Difficult Long-Context Task Design}
\label{sec:next}

\paragraph{Challenges.} Designing meaningful long-context tasks amidst rapid model progress is profoundly challenging, making the deficiency in tasks representing difficulty on both the \textit{dispersion} and \textit{scope} axes unsurprising. 
One source of this challenge is the lack of diverse, coherent long texts, as models' context windows can now be comparable to 
the length of the New Testament\footnote{\url{www.readinglength.com/book/isbn-0190909005}}
and the Odyssey.\footnote{\url{www.readinglength.com/book/isbn-0140268863}}
The methodologies discussed in \S\ref{sec:survey} for creating long context tasks -- lengthening short context tasks and synthetically creating length-adjustable tasks -- are preferred for their straightforward definition and the incremental adjustments they require for existing data. 
They rely on the common understanding of machine comprehension as formulated with short context in mind \cite{dunietz-etal-2020-test}, and therefore they are intuitive and easy to comprehend for NLP researchers without domain expertise (e.g., in law or biomedical fields that have long contexts). 

\paragraph{Future work.} 
The goals laid forward in this work are clear: For more durable and robust measurement of long-context capabilities, we must design tasks that explicitly target both the \textit{dispersion} and \textit{scope} capabilities of models. 
How can this be achieved? As mentioned, one possible avenue is to rely more on \textit{tasks that require domain expertise}, such as legal documents~\cite{bruno2022lawngnli}, financial reports~\cite{reddy2024docfinqa}, biomedical publications~\cite{stylianou2021improved}, and so on. In specialized domains, it is common that \textit{dispersion} will be naturally higher~\cite{zhao-etal-2022-multihiertt}. Tasks that involve \textit{implicit aggregations over structured data}, such as table manipulation~\cite{caciularu2024tact}, are possible avenues for increasing both scope and dispersion synthetically by leveraging the data structure. 
In this work, we argue that an \textit{explicit vocabulary} for such properties of difficulty is what can enable more informed techniques to design difficult tasks in the future.

\section{Conclusions}
We present a taxonomy of factors that make long-context tasks more challenging compared to short ones. This is in contrast with the existing literature that refers only to the length of the input as the hallmark of long context, and as a result ends up conflating tasks of different character when assessing the ability of models to understand longer text. 
We reviewed works on evaluation in a long-context setting and found that the most challenging setting, in which the information needed is of large \textit{scope} and is highly \textit{dispersed} within the input, is under-explored. 
Finally, we suggested some leads for future work to tackle this imbalance towards a more informative long context evaluation.

\section{Limitations}
\label{sec:limitations}

\paragraph{Formality.} In the context of this work, we have deliberately adhered to a taxonomy based on an intuitive description, in the interest of utility to a wide diversity of research and flexibility for future work. Difficulty in searching for and extracting information, and quantity of information, are both vague terms that can only be grounded in the context of a specific family of tasks and use-cases. We intend for this work to serve as a call to action and a tool for a shared vocabulary in the interest of more informed long-context task design in the future, rather than to anchor the taxonomy to a specific and fragile point in time.

\paragraph{Retrieval is still interesting.} 
Although we argue that small scope and low dispersion tasks are the least indicative of the model's ability to long-context capabilities, tasks that are well-served by implicit retrieval or by traditional retrieval-based pipelines are certainly relevant and useful in a variety of common use-cases~\cite{stylianou2021improved, bruno2022lawngnli, gao2023rarr}.

\paragraph{Other uses for a long-context window.} 
This paper deals only with long inputs that serve as inputs to a task. The long context of course can have other purposes as well, like containing many in-context examples \cite{bertsch2024incontext} or containing other modalities and structures \cite{jiang-etal-2023-structgpt}.

\section*{Acknowledgments}
The authors would like to thank Gabriel Stanovsky for the fruitful discussions. This work has been funded by the Israel Science Foundation, grant number 23/670, for which we are grateful.

\bibliography{custom,anthology,anthology2}

\appendix

\section{Benchmark Scope-Dispersion Classification}
\label{sec:benchmark_scope_dispersion_classification}
In \autoref{tab:tasks_scope_dispersion_classification} we delineate the different long-context benchmarks, as well as classify them according to how challenging they are scope-wise and dispersion-wise.

\begin{table*}[]
\centering
\resizebox{0.75\textwidth}{!}{%
\begin{tabular}{c|l|l|}
\cline{2-3}
                                                                                                             & \multicolumn{1}{c|}{LOW SCOPE}                                                                 & \multicolumn{1}{c|}{HIGH SCOPE}                                                                         \\ \hline
\multicolumn{1}{|c|}{}                                                                                       & QA                                                                                             & QBS                                                                                                     \\
\multicolumn{1}{|c|}{}                                                                                       & \hspace{2mm} {\small Qasper \citep{dasigi-etal-2021-dataset}}                                  & \hspace{2mm} {\small QMSum \citep{zhong-etal-2021-qmsum}}                                               \\
\multicolumn{1}{|c|}{}                                                                                       & \hspace{2mm} {\small NarrativeQA \citep{narrativeQA}}                                          & \hspace{2mm} {\small SQuALITY \citep{wang-etal-2022-squality}}                                          \\
\multicolumn{1}{|c|}{}                                                                                       & \hspace{2mm} {\small Short-dependency QA \citep{li2023loogle}}                                 & \hspace{2mm} {\small Related Work Summarization \citep{an2023leval}}                                    \\
\multicolumn{1}{|c|}{}                                                                                       & \hspace{2mm} {\small MultiFieldQA \citep{bai2023longbench}}                                    & \hspace{2mm} {\small SPACE \citep{angelidis-etal-2021-extractive}}                                      \\
\multicolumn{1}{|c|}{}                                                                                       & \hspace{2mm} {\small LitM (QA) \citep{liu2024lost}}                                            & \hspace{2mm} {\small WebBrain-G \citep{qian2023webbrainlearninggeneratefactually}}                      \\
\multicolumn{1}{|c|}{}                                                                                       & \hspace{2mm} {\small L-eval (MC QA) \citep{an2023leval}}                                       & \hspace{2mm} {\small AquaMuse \citep{kulkarni2020aquamuseautomaticallygeneratingdatasets}}              \\
\multicolumn{1}{|c|}{}                                                                                       & \hspace{2mm} {\small NQ \citep{kwiatkowski-etal-2019-natural}}                                 & \hspace{2mm} {\small FINDSum-Liquidity \citep{liu2023longtextmultitablesummarization}}                  \\
\multicolumn{1}{|c|}{}                                                                                       & \hspace{2mm} {\small RULER (single-hop QA) \citep{hsieh2024ruler}}                             & Aggregation                                                                                             \\
\multicolumn{1}{|c|}{}                                                                                       & \hspace{2mm} {\small MeetingQA \citep{prasad-etal-2023-meetingqa}}                             & \hspace{2mm} {\small ZeroSCROLLS (SpaceDigest \& BookSumSort) \citep{shaham-etal-2023-zeroscrolls}}     \\
\multicolumn{1}{|c|}{}                                                                                       & \hspace{2mm} {\small BABILong (tasks 1,4-6,9-10) \citep{kuratov2024searchneedles11mhaystack}}  & \hspace{2mm} {\small PassageCount \citep{bai2023longbench}}                                             \\
\multicolumn{1}{|c|}{}                                                                                       & \hspace{2mm} {\small Giraffe (2 tasks) \citep{pal2023giraffeadventuresexpandingcontext}}       & \hspace{2mm} {\small FINDSum-ROO \citep{liu2023longtextmultitablesummarization}}                        \\
\multicolumn{1}{|c|}{}                                                                                       & Retrieval                                                                                      & Aspect-based Summarization                                                                              \\
\multicolumn{1}{|c|}{}                                                                                       & \hspace{2mm} {\small LitM (Key-value Retrieval) \citep{liu2024lost}}                           & \hspace{2mm} {\small ACLSum \citep{takeshita-etal-2024-aclsum}}                                         \\
\multicolumn{1}{|c|}{}                                                                                       & \hspace{2mm} {\small MultiDoc2Dial (GSP) \citep{feng-etal-2021-multidoc2dial}}                 & \hspace{2mm} {\small OpenAsp \citep{amar-etal-2023-openasp}}                                            \\
\multicolumn{1}{|c|}{}                                                                                       & \hspace{2mm} {\small TopicRet \citep{longchat2023}}                                            & Text Sorting                                                                                            \\
\multicolumn{1}{|c|}{}                                                                                       & \hspace{2mm} {\small Wiki-GenBen \citep{zhang2024retrievalbasedfulllengthwikipediageneration}} & \hspace{2mm} {\small Bamboo (ShowsSort \& ReportSumSort) \citep{dong-etal-2024-bamboo}}                 \\
\multicolumn{1}{|c|}{}                                                                                       & \hspace{2mm} {\small RULER (S-NIAH \& MK-NIAH) \citep{hsieh2024ruler}}                         & Retrieval                                                                                               \\
\multicolumn{1}{|c|}{}                                                                                       & \hspace{2mm} {\small LongChat-Lines \citep{pal2023giraffeadventuresexpandingcontext}}          & \hspace{2mm} {\small PassageRetrieval \citep{bai2023longbench}}                                         \\
\multicolumn{1}{|c|}{}                                                                                       & NLI                                                                                            & LFQA                                                                                                    \\
\multicolumn{1}{|c|}{}                                                                                       & \hspace{2mm} {\small LawngNLI \citep{bruno2022lawngnli}}                                       & \hspace{2mm} {\small LongFQA \citep{an2023leval}}                                                       \\
\multicolumn{1}{|c|}{}                                                                                       & \hspace{2mm} {\small ContractNLI \citep{koreeda2021contractnlidatasetdocumentlevelnatural}}    & NLI                                                                                                     \\
\multicolumn{1}{|c|}{}                                                                                       & \hspace{2mm} {\small Hallucination Detection \citep{dong-etal-2024-bamboo}}                    & \hspace{2mm} {\small Legal Case Entailment \citep{nguyen2024captain}}                                   \\
\multicolumn{1}{|c|}{}                                                                                       & {\color[HTML]{333333} \hspace{2mm} {\small FLenQA (3 tasks) \citep{levy2024task}}}             &                                                                                                         \\
\multicolumn{1}{|c|}{}                                                                                       & Fill-mask                                                                                      &                                                                                                         \\
\multicolumn{1}{|c|}{}                                                                                       & \hspace{2mm} {\small Cloze \citep{li2023loogle}}                                               &                                                                                                         \\
\multicolumn{1}{|c|}{}                                                                                       & NLG                                                                                            &                                                                                                         \\
\multicolumn{1}{|c|}{\multirow{-28}{*}{\rotatebox[origin=c]{90}{\normalfont \hspace{-4mm} LOW DISPERSION}}}  & \hspace{2mm} {\small MultiDoc2Dial (ARG) \citep{feng-etal-2021-multidoc2dial}}                 &                                                                                                         \\ \hline
\multicolumn{1}{|c|}{}                                                                                       & QA                                                                                             & Summarization                                                                                           \\
\multicolumn{1}{|c|}{}                                                                                       & \hspace{2mm} {\small QuALITY \citep{pang-etal-2022-quality}}                                   & \hspace{2mm} {\small GovReport \citep{huang2021efficientattentionslongdocument}}                        \\
\multicolumn{1}{|c|}{}                                                                                       & \hspace{2mm} {\small Long-dependency QA \citep{li2023loogle}}                                  & \hspace{2mm} {\small SummScreenFD \citep{chen2022summscreendatasetabstractivescreenplay}}               \\
\multicolumn{1}{|c|}{}                                                                                       & \hspace{2mm} {\small DuReader \citep{bai2023longbench}}                                        & \hspace{2mm} {\small Loogle (Summarization) \citep{li2023loogle}}                                       \\
\multicolumn{1}{|c|}{}                                                                                       & \hspace{2mm} {\small SFcition QA \citep{an2023leval}}                                          & \hspace{2mm} {\small VCSUM \citep{bai2023longbench}}                                                    \\
\multicolumn{1}{|c|}{}                                                                                       & \hspace{2mm} {\small ExpertQA \citep{malaviya-etal-2024-expertqa}}                             & \hspace{2mm} {\small Self-critiquing \citep{saunders2022selfcritiquing}}                                \\
\multicolumn{1}{|c|}{}                                                                                       & \hspace{2mm} {\small DocFinQA \citep{reddy2024docfinqa}}                                       & \hspace{2mm} {\small Abstract Generation \citep{an2023leval}}                                           \\
\multicolumn{1}{|c|}{}                                                                                       & \hspace{2mm} {\small BABILong (tasks 2-3,12) \citep{kuratov2024searchneedles11mhaystack}}      & \hspace{2mm} {\small Multi-News \citep{fabbri2019multinews}}                                            \\
\multicolumn{1}{|c|}{}                                                                                       & \hspace{2mm} {\small Bamboo (QA) \citep{dong-etal-2024-bamboo}}                                & \hspace{2mm} {\small BigPatent \citep{sharma-etal-2019-bigpatent}}                                      \\
\multicolumn{1}{|c|}{}                                                                                       & Multi-hop QA                                                                                   & \hspace{2mm} {\small Scientific Summarization \citep{cohan2018discourseawareattentionmodelabstractive}} \\
\multicolumn{1}{|c|}{}                                                                                       & \hspace{2mm} {\small MuSiQue \citep{trivedi2022musique}}                                       & \hspace{2mm} {\small BillSum \citep{kornilova-eidelman-2019-billsum}}                                   \\
\multicolumn{1}{|c|}{}                                                                                       & \hspace{2mm} {\small HotpotQA \citep{yang2018hotpotqa}}                                        & \hspace{2mm} {\small HowSumm \citep{boni2021howsummmultidocumentsummarizationdataset}}                  \\
\multicolumn{1}{|c|}{}                                                                                       & \hspace{2mm} {\small Multi-hop Tracing \citep{hsieh2024ruler}}                                 & \hspace{2mm} {\small ODSum \citep{zhou2023odsumnewbenchmarksopen}}                                      \\
\multicolumn{1}{|c|}{}                                                                                       & \hspace{2mm} {\small RULER (multi-hop QA) \citep{hsieh2024ruler}}                              & \hspace{2mm} {\small Klexikon (Summarization) \citep{aumiller-gertz-2022-klexikon}}                     \\
\multicolumn{1}{|c|}{}                                                                                       & \hspace{2mm} {\small 2WikiMultihopQA \citep{ho-etal-2020-constructing}}                        & \hspace{2mm} {\small Booksum \citep{kryściński2022booksumcollectiondatasetslongform}}                   \\
\multicolumn{1}{|c|}{}                                                                                       & NLI                                                                                            & \hspace{2mm} {\small MeetingBank \citep{hu-etal-2023-meetingbank}}                                      \\
\multicolumn{1}{|c|}{}                                                                                       & \hspace{2mm} {\small FLenQA (3 rand. placement tasks) \citep{levy2024task}}                    & Text Simplification                                                                                     \\
\multicolumn{1}{|c|}{}                                                                                       & \hspace{2mm} {\small Legal Textual Entailment \citep{nguyen2024captain}}                       & \hspace{2mm} {\small Klexikon (Simplification) \citep{aumiller-gertz-2022-klexikon}}                    \\
\multicolumn{1}{|c|}{}                                                                                       & Code Understanding                                                                             & Reasoning                                                                                               \\
\multicolumn{1}{|c|}{}                                                                                       & \hspace{2mm} {\small LCC \citep{guo2023longcoderlongrangepretrainedlanguage}}                  & \hspace{2mm} {\small Long ListOps \citep{tay2020long}}                                                  \\
\multicolumn{1}{|c|}{}                                                                                       & \hspace{2mm} {\small RepoBench-P \citep{liu2023repobenchbenchmarkingrepositorylevelcode}}      & Retrieval                                                                                               \\
\multicolumn{1}{|c|}{}                                                                                       & \hspace{2mm} {\small CodeU \citep{an2023leval}}                                                & \hspace{2mm} {\small LRA (task 3) \citep{tay2020long}}                                                  \\
\multicolumn{1}{|c|}{}                                                                                       & \hspace{2mm} {\small PrivateEval \citep{dong-etal-2024-bamboo}}                                &                                                                                                         \\
\multicolumn{1}{|c|}{}                                                                                       & Classification                                                                                 &                                                                                                         \\
\multicolumn{1}{|c|}{}                                                                                       & \hspace{2mm} {\small LRA (tasks 2, 4-6) \citep{tay2020long}}                                   &                                                                                                         \\
\multicolumn{1}{|c|}{}                                                                                       & Retrieval                                                                                      &                                                                                                         \\
\multicolumn{1}{|c|}{}                                                                                       & \hspace{2mm} {\small COLIEE (tasks 1,3,4) \citep{nguyen2024captain}}                           &                                                                                                         \\
\multicolumn{1}{|c|}{}                                                                                       & \hspace{2mm} {\small RULER (MV-NIAH \& MQ-NIAH) \citep{hsieh2024ruler}}                        &                                                                                                         \\
\multicolumn{1}{|c|}{}                                                                                       & Next Token Prediction                                                                          &                                                                                                         \\
\multicolumn{1}{|c|}{}                                                                                       & \hspace{2mm} {\small PG-19 \citep{rae2019compressivetransformerslongrangesequence}}            &                                                                                                         \\
\multicolumn{1}{|c|}{}                                                                                       & \hspace{2mm} {\small Bamboo (LM) \citep{dong-etal-2024-bamboo}}                                &                                                                                                         \\
\multicolumn{1}{|c|}{}                                                                                       & Reasoning                                                                                      &                                                                                                         \\
\multicolumn{1}{|c|}{}                                                                                       & \hspace{2mm} {\small DocMath-Eval \citep{Zhao2023DocMathEvalEN}}                               &                                                                                                         \\
\multicolumn{1}{|c|}{}                                                                                       & \hspace{2mm} {\small BABILong (tasks 14-20) \citep{kuratov2024searchneedles11mhaystack}}       &                                                                                                         \\
\multicolumn{1}{|c|}{}                                                                                       & Aggregation                                                                                    &                                                                                                         \\
\multicolumn{1}{|c|}{}                                                                                       & \hspace{2mm} {\small RULER (2 Aggr. tasks) \citep{hsieh2024ruler}}                             &                                                                                                         \\
\multicolumn{1}{|c|}{}                                                                                       & \hspace{2mm} {\small BABILong (tasks 7-8) \citep{kuratov2024searchneedles11mhaystack}}         &                                                                                                         \\
\multicolumn{1}{|c|}{}                                                                                       & NLU                                                                                            &                                                                                                         \\
\multicolumn{1}{|c|}{}                                                                                       & \hspace{2mm} {\small Academic Feedback Generation \citep{an2023leval}}                         &                                                                                                         \\
\multicolumn{1}{|c|}{}                                                                                       & \hspace{2mm} {\small CUAD \citep{hendrycks2021cuadexpertannotatednlpdataset}}                  &                                                                                                         \\
\multicolumn{1}{|c|}{}                                                                                       & Factuality Evaluation                                                                          &                                                                                                         \\
\multicolumn{1}{|c|}{}                                                                                       & \hspace{2mm} {\small LongSciVerify \citep{bishop2024longdocfactscore}}                         &                                                                                                         \\
\multicolumn{1}{|c|}{}                                                                                       & Coreference Resolution                                                                         &                                                                                                         \\
\multicolumn{1}{|c|}{\multirow{-44}{*}{\rotatebox[origin=c]{90}{\normalfont \hspace{-4mm} HIGH DISPERSION}}} & \hspace{2mm} {\small BABILong (tasks 11,13) \citep{kuratov2024searchneedles11mhaystack}}       &                                                                                                         \\ \hline
\end{tabular}%
}
\caption{Classification of long-context benchmarks in terms of Scope and Dispersion.}
\label{tab:tasks_scope_dispersion_classification}
\end{table*}

\end{document}